\documentclass[letterpaper, 10 pt, conference]{ieeeconf}  

\IEEEoverridecommandlockouts                              

\usepackage{graphicx}

\title{\LARGE \bf
EVOLVE: Emotion and Visual Output Learning via LLM Evaluation
}

\author{Jordan Sinclair$^{1}$ and Christopher Reardon$^{1}$
\thanks{$^{1}$Department of Computer Science, Ritchie School of Computer Science and Engineering, University of Denver, USA.
        {\tt\small \{jordan.sinclair,christopher.reardon\}@du.edu}}%
}

\begin{document}

\maketitle
\thispagestyle{empty}
\pagestyle{empty}

\begin{abstract}

Human acceptance of social robots is greatly effected by empathy and perceived understanding. This necessitates accurate and flexible responses to various input data from the user. While systems such as this can become increasingly complex as more states or response types are included, new research in the application of large language models towards human-robot interaction has allowed for more streamlined perception and reaction pipelines. LLM-selected actions and emotional expressions can help reinforce the realism of displayed empathy and allow for improved communication between the robot and user. Beyond portraying empathy in spoken or written responses, this shows the possibilities of using LLMs in actuated, real world scenarios. In this work we extend research in LLM-driven nonverbal behavior for social robots by considering more open-ended emotional response selection leveraging new advances in vision-language models, along with emotionally aligned motion and color pattern selections that strengthen conveyance of meaning and empathy. 

\end{abstract}

\section{INTRODUCTION}

Many social robotics applications necessitate high communication standards, including both verbal and nonverbal conveyance. New approaches utilizing Large Language Models (LLMs) have shown promise in alleviating the design of complex interaction systems and providing more adaptable interplay between human and robot agents. Work in this area has sought to develop more robust written and spoken communication in social robotics, as well as to align nonverbal behavior with verbal messages.

While the ability to effectively communicate and retain user attention for longer periods of time is important in many HRI settings, eliciting an impression of empathy through nonverbal behavior can be critical to acceptance of and trust in social robots \cite{GoodInteracting}. Through a comprehensive survey over several LLM-based actions, \cite{Understanding} discovered that social robots elicited higher expectations for more nuanced nonverbal cues including a breadth of behavior types.
Conveying affects that are aligned with the user's emotional state can be critical in building trust around experienced empathy and personalization from a social robot \cite{irfan2024recommendations}. 
Multi-modal feedback have profound impacts on successful empathetic interaction, as notions inferred from robot actions can be understood much easier with systematic actions taken in alignment with an emotional response \cite{Understanding, NonverbalCues}. 

\begin{figure}[t]
    \centering
    \includegraphics[width=3in]{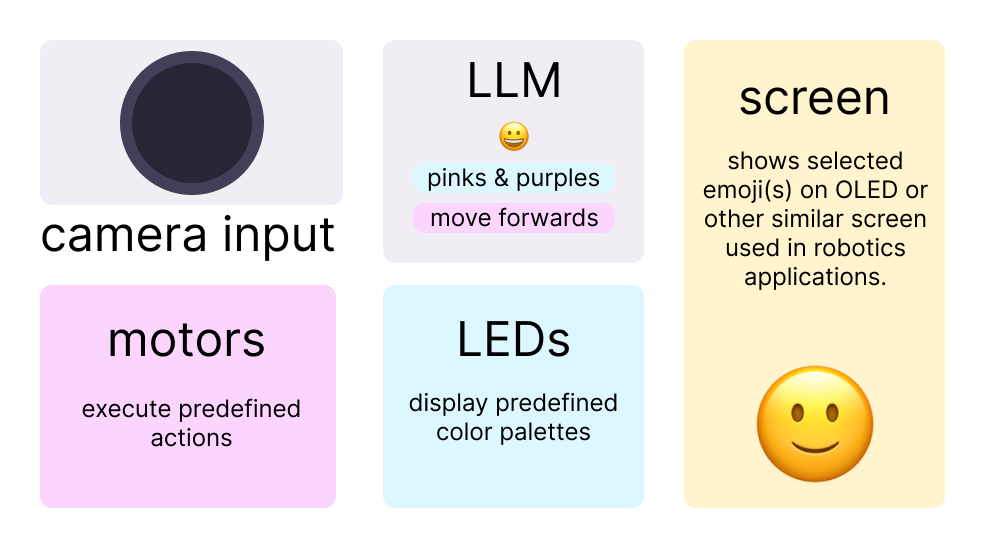}
    \includegraphics[width=1.5in]{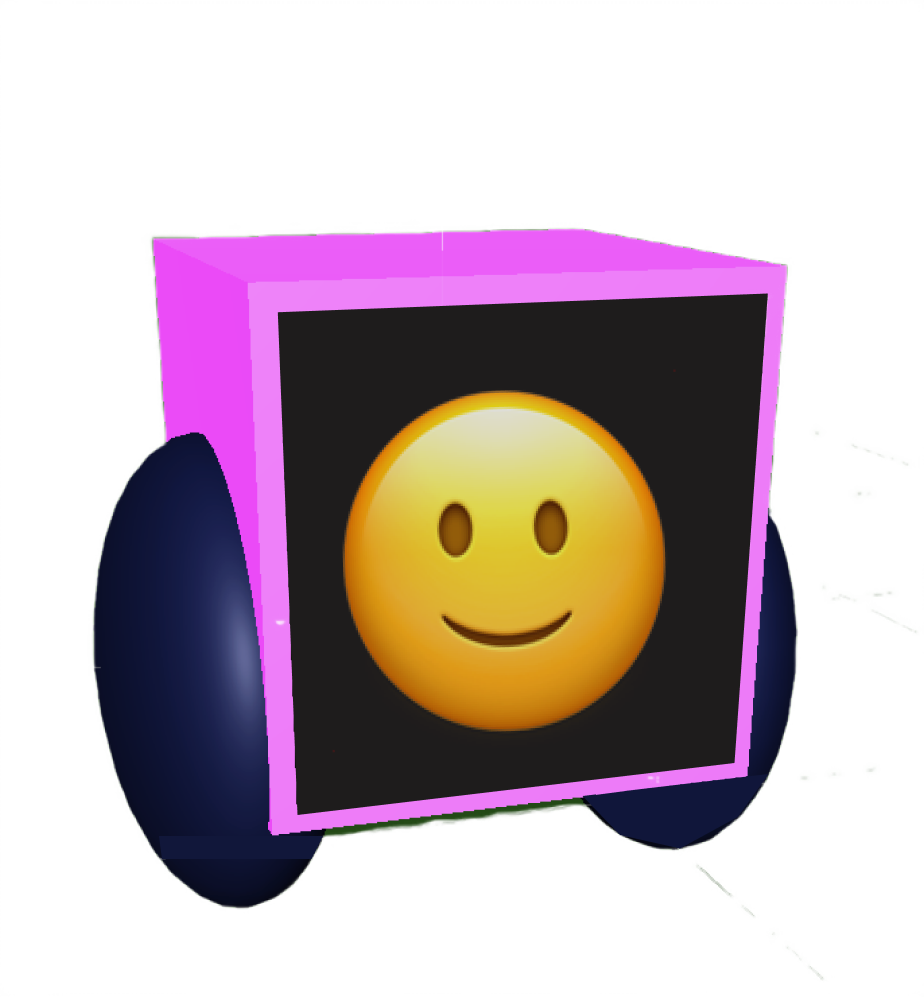}
    \caption{(a) The LLM evaluates a camera image input and determines three visual outputs that evolve with new data: an emoji representing affective response, a color palette to be visualized on LEDs, and a motion pattern. (b) Potential robot design with these characteristics.}
    \label{fig:system}
\end{figure}

Atomic actions, or sub-modules of larger behavior patterns that an LLM can choose from, can enhance the variability in emotional responses and motifs \cite{lami}. This allows for a more seamless integration of verbal and nonverbal behaviors, and ultimately more complex interaction capabilities for social robots. Additionally, this kind of subdivided action schema can be used to evaluate many attributes towards promoting empathetic responses, including tone of voice, nonverbal cues, and facial expressions \cite{lee2023developingsocialrobotsempathetic}.
However, atomic actions with limited sentiments might not be sufficient to accommodate complex emotion in the user. This work investigates the possibility of a more open-ended response selection by leveraging an LLM's internal domain knowledge of emojis and other affective imagery capable of representing emotional states. We also employ recent advances in vision-language models with an image or camera input, as suggested in \cite{Understanding} and \cite{NonverbalCues}. Additionally, we evaluate both motion and color \cite{color} pattern elicitation through atomic action selection \cite{lami, lee2023developingsocialrobotsempathetic}. We selected these decision categories based on a theoretical robot design that could contain an LED strip visible through the robot's clear acrylic casing, two wheels, and an OLED screen to display an emoji. 
To summarize, the contributions of this paper include: 

\begin{itemize}
    \item Utilizing atomic actions \cite{lami} towards selecting motion and color behaviors aligned with expected empathetic responses. 
    \item Introducing a novel approach integrating vision-language models for camera input, allowing more streamlined nonverbal interaction with a visual emphasis. 
    \item Investigating a more open-ended approach to empathetic response selection using a larger sample space consisting of emojis, knowledge of which stems from initial LLM training data.
\end{itemize}

A system diagram is shown in Fig. \ref{fig:system}. 

\section{Prompting \& Initial Results}

\begin{figure}[t]
    \centering
    \includegraphics[width=3.5in]{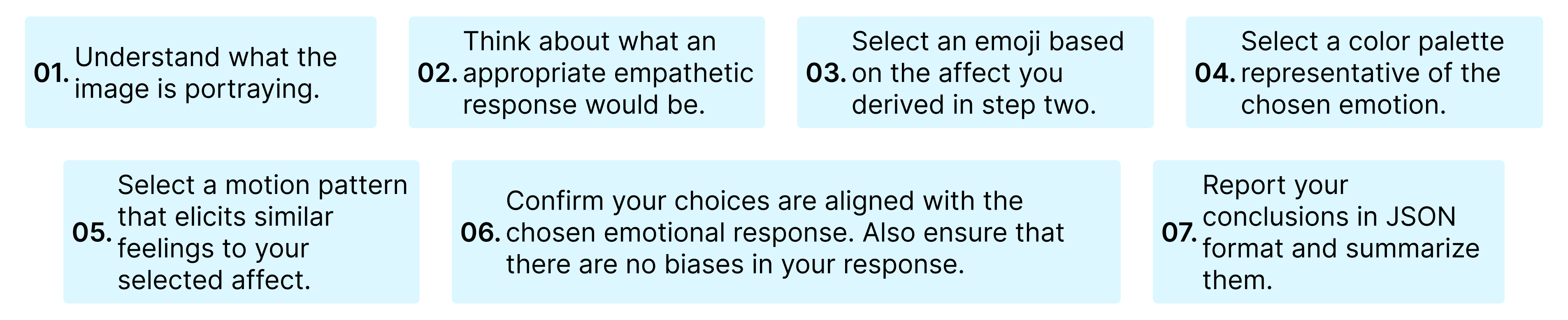}
    \caption{Prompt procedure}
    \label{fig:prompts}
\end{figure}

We crafted a prompt for the LLM characterizing the goals of the procedure in seven steps, see Fig. \ref{fig:prompts}, following several general prompting techniques, including giving the model more time to interpret the questions (steps one and two), leveraging in-domain knowledge of emojis (step three), using clear delimiters (steps four and five), verification of results (step six), and using a structured output (step seven). 
The model produces an emoji representing the emotional response to the human in the given image, a motion pattern selected from a predefined list of options, a color palette (meant to represent a sequence of LEDs), and a short explanation of the chosen attributes. 

\begin{figure}[t]
    \centering
    \includegraphics[width=3in]{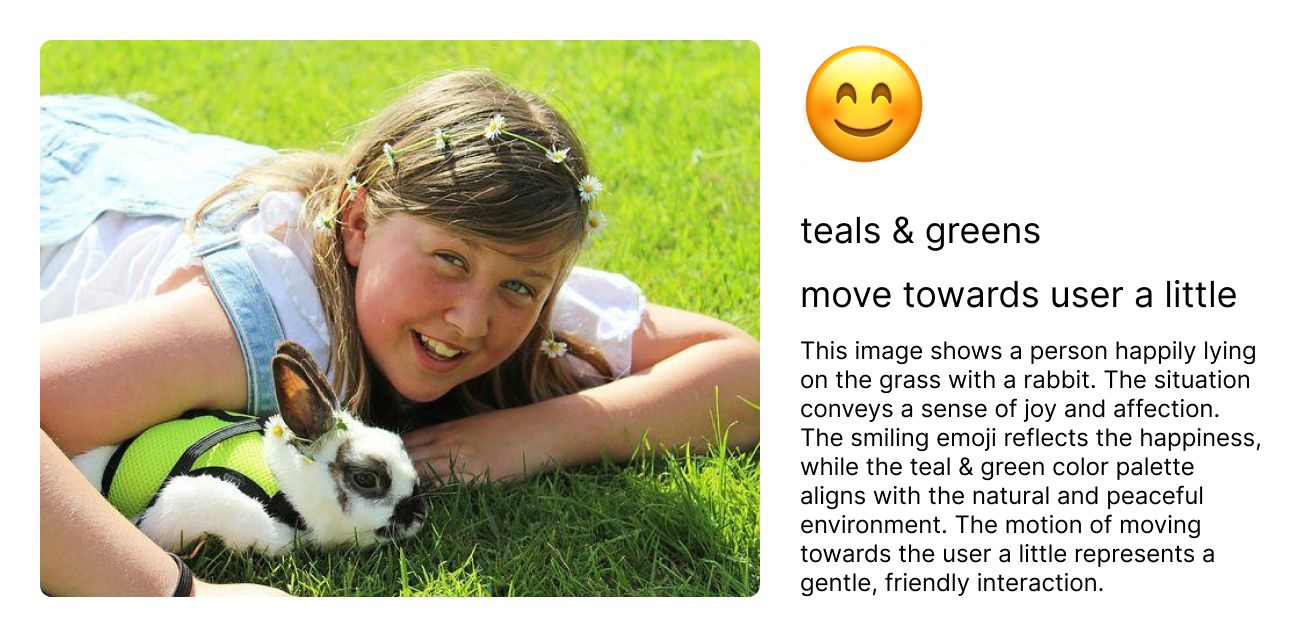}
    \caption{LLM response for image initially labelled as contentment.}
    \label{fig:contentment}
\end{figure}

\begin{figure}[t]
    \centering
    \includegraphics[width=3in]{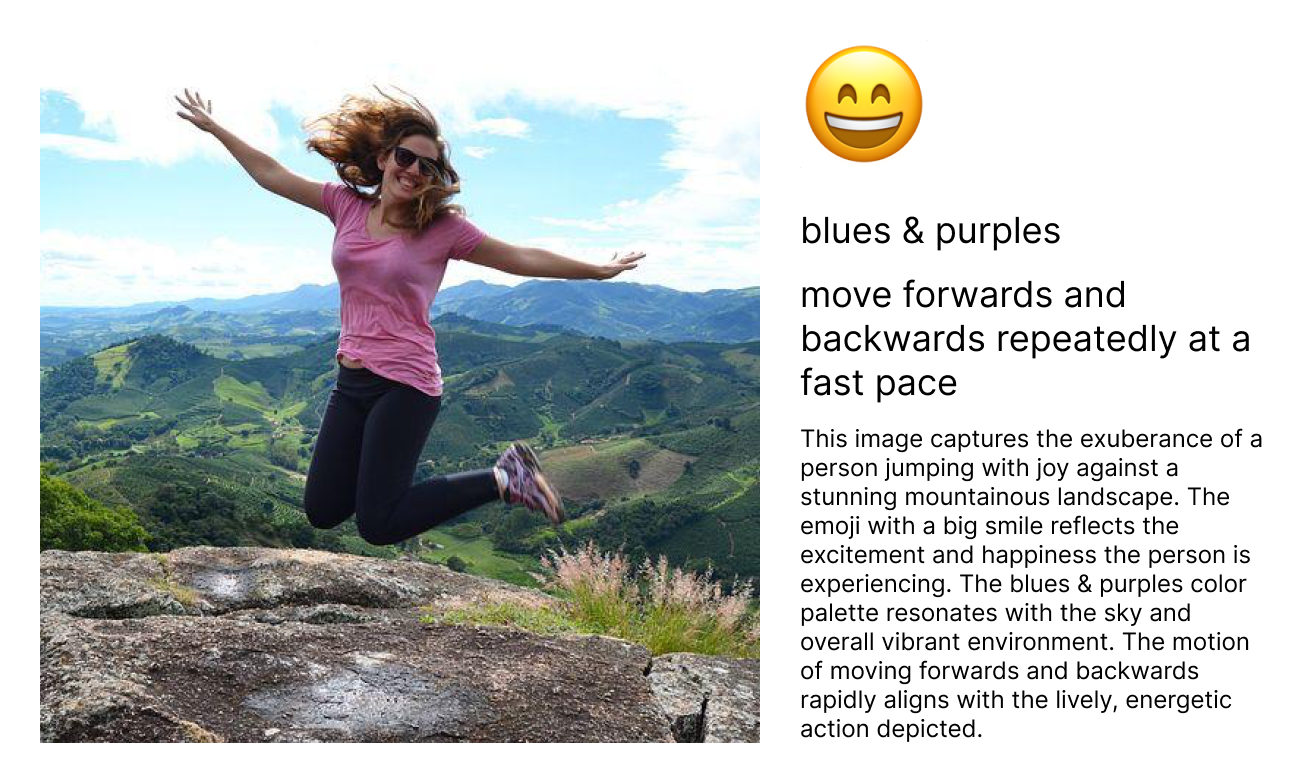}
    \caption{LLM response for image initially labelled as excitement.}
    \label{fig:excitement}
\end{figure}

\begin{figure}[t!]
    \centering
    \includegraphics[width=3in]{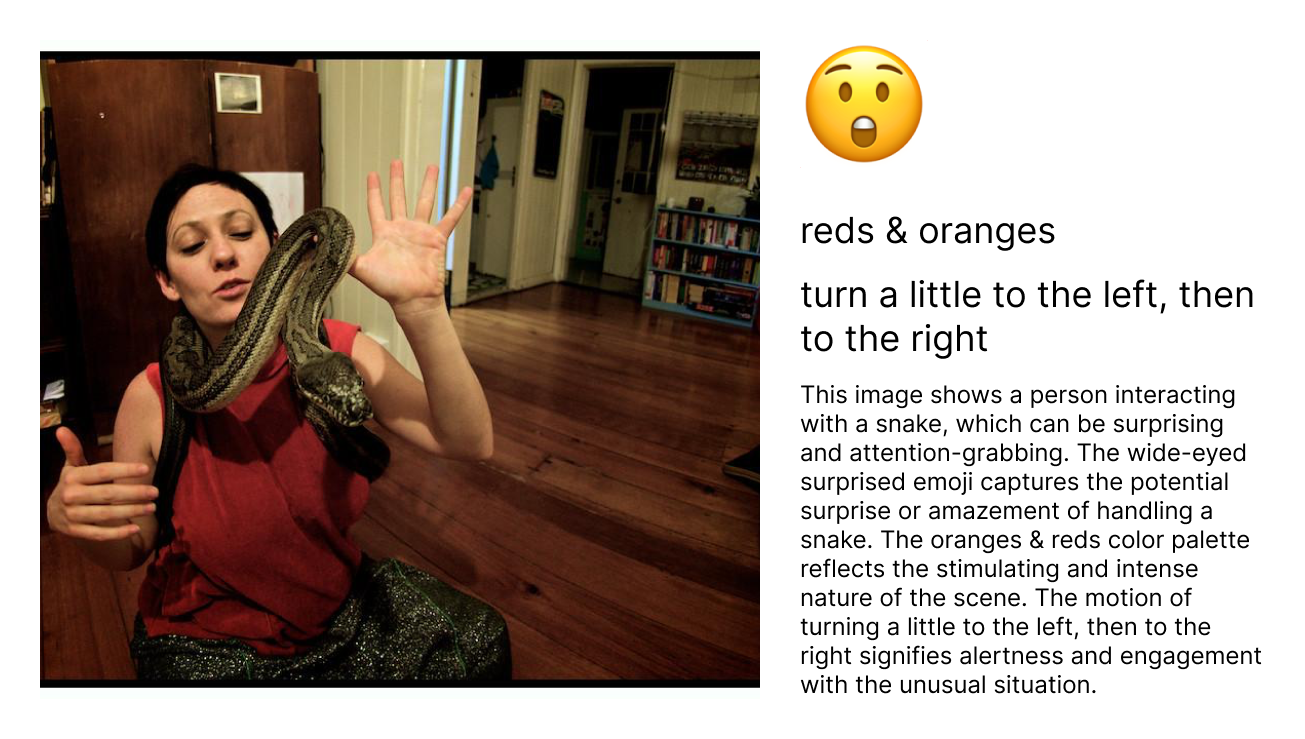}
    \caption{LLM response for image initially labelled as fear.}
    \label{fig:fear}
\end{figure}

We validate our initial work using input images selected from EmoSet \cite{emoset}, which provides data labelled with emotional affects that we used as a baseline comparison against the LLM responses. 
Fig. \ref{fig:contentment} was initially labelled in the EmoSet database \cite{emoset} with an affect of contentment, which would seem to align with the models interpretation leaning towards greens and blues, and moving towards the user. Fig \ref{fig:excitement} seemed to align fairly well with the intended label of excitement, however the LLM appeared to pull the color palette from the image itself rather than selecting one that aligned with a desired emotional response. In future work we will focus on this by adjusting the prompt further and exploring alternative inputs such as black and white photos, which might lead to results focused more on empathetic conveyance and less on the image itself. The LLM seemed to have a reasonable interpretation of Fig. \ref{fig:fear}, representative of fear, as it selected red and orange colors to communicate danger. The motion pattern could be seen as representing worry and concern. While the selections were not perfectly aligned with the input images in all cases, this initial work signifies the potential of visual input as a medium for empathetic response understanding in social robots. 

\section{Conclusion}

We presented a novel contribution to the body of work focused on LLM integration with empathy modeling in social robots. Specifically, we explored the possibility and efficacy of using vision-language models for image analysis towards emotional output realization. We also developed a promising paradigm for more open-ended empathetic response selection utilizing internal domain knowledge, while simultaneously using action building blocks to create complex and interactive behaviors related to motion and color patterns. Finally, we presented an initial concept for a social robot utilizing these features. Our initial results showed promise in generally aligning with expected affects, although more work is needed in determining how color and motion preferences differ between users and what bias exists in the model itself, which could contribute to misrepresentation in responses, and ultimately to mistrust in the agent. We also plan to integrate our proposed methodology with a Retrieval Augmented Generation (RAG) pipeline towards a memory of user interactions that could be used for targeted personalization and evolution in robot behavior. This could help overcome some model bias by remembering interactions that the user responded positively to.

\bibliographystyle{./IEEEtran}
\bibliography{IEEEbcpat}

\end{document}